\documentclass{egpubl}
\usepackage{egsgp19}
 
 \SpecialIssuePaper         

\usepackage[T1]{fontenc}
\usepackage{dfadobe}  

\usepackage{cite}  
\BibtexOrBiblatex
\electronicVersion
\PrintedOrElectronic
\ifpdf \usepackage[pdftex]{graphicx} \pdfcompresslevel=9
\else \usepackage[dvips]{graphicx} \fi

\usepackage{egweblnk} 
\usepackage{todonotes}
\usepackage{amsmath}
\usepackage{amssymb}
\usepackage{bbold}
\usepackage{siunitx}
\usepackage{layouts}
\usepackage{overpic}
\usepackage{wrapfig}
\usepackage[export]{adjustbox}
\usepackage{rotating}
\usepackage{color,soul}
\graphicspath{{figures/}}
\newcommand{\norm}[1]{\left\lVert#1\right\rVert}

\title[A Convolutional Decoder for Point Clouds]%
      {A Convolutional Decoder for Point Clouds\\ using Adaptive Instance Normalization}

\author[I. Lim, M. Ibing, L. Kobbelt]
{\parbox{\textwidth}{\centering\hspace{\stretch{2}}Isaak Lim\thanks{Equal Contribution}\hspace{\stretch{1}}
        Moritz Ibing\footnotemark[1]\hspace{\stretch{1}}
        Leif Kobbelt\hspace{\stretch{2}}
        }
        \\
{\parbox{\textwidth}{\centering Visual Computing Institute, RWTH Aachen University
       }
}
}

\begin{document}

\teaser{
\begin{overpic}[width=\linewidth]{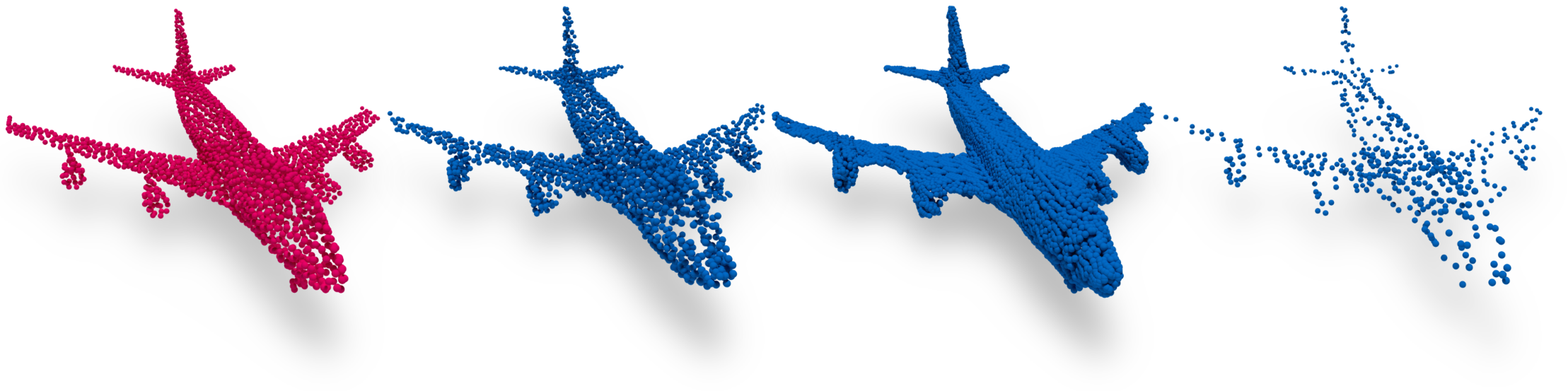}
\put(15.5,20){$2500$ points}
\put(40.5,20){$2500$ points}
\put(65.5,20){$15000$ points}
\put(91,20){$500$ points}
\end{overpic}
 \centering
 \caption{We show decoding results (blue) for an input shape (red) from the test set.
 Our convolutional autoencoder with Adaptive Instance Normalization was trained to output 2500 points for inputs with 2500 points.
 We also visualize outputs from our decoder with a much higher (15000) or lower (500) number of points than the number used during training.
 Note that with 15000 points we are able to robustly and densely sample the underlying geometry of the input point cloud.
 Conversely, with 500 points our method is still able to capture the overall shape of the original input.
 }
 \label{fig:teaser}
}

\maketitle
\begin{abstract}
   Automatic synthesis of high quality 3D shapes is an ongoing and challenging area of research.
   While several data-driven methods have been proposed that make use of neural networks to generate 3D shapes, none of them reach the level of quality that deep learning synthesis approaches for images provide.
   In this work we present a method for a convolutional point cloud decoder/generator that makes use of recent advances in the domain of image synthesis.
   Namely, we use Adaptive Instance Normalization and offer an intuition on why it can improve training.
   Furthermore, we propose extensions to the minimization of the commonly used Chamfer distance for auto-encoding point clouds.
   In addition, we show that careful sampling is important both for the input geometry and in our point cloud generation process to improve results.
   The results are evaluated in an auto-encoding setup to offer both qualitative and quantitative analysis.
   The proposed decoder is validated by an extensive ablation study and is able to outperform current state of the art results in a number of experiments.
   We show the applicability of our method in the fields of point cloud upsampling, single view reconstruction, and shape synthesis.


\end{abstract}  
\section{Introduction}
The question of how to represent 3D geometry as input for neural networks is still an ongoing field of research.
Most recent papers (e.g.~\cite{qi2017pointnet, qi2017pointnet++,atzmon2018pcnn,fey2018splinecnn,li2018pointcnn}) focus on how to encode the input in a manner such that its latent representation can then be used for tasks such as classification or segmentation.
However, a smaller amount of work has been done on how high-fidelity 3D shapes can be generated by a decoder/generator network.
We investigate the problem of generating 3D shapes in an auto-encoding setup.
This allows us to evaluate results both qualitatively and quantitatively.
While a number of previous works focus on the encoder, we mainly target the decoder/generator in this paper.

Synthesis of 3D shapes is a time consuming task (especially for non-expert users), which is why a number of data-driven approaches have been proposed to tackle this problem.
Methods range from combining parts of a shape collection to create novel configurations over deformation based approaches to the full synthesis of voxelized, meshed or point sampled 3D shapes.
While impressive results have been presented, generated 3D shapes have not yet reached a quality that is comparable to the state of the art in image generation, such as recently presented by Karras et al.~\cite{adain18}.

We are interested in the complete synthesis of 3D shapes. In particular we investigate the generation of 3D point clouds since voxelized representation incur a heavy memory cost.
At the same time we want to benefit from recent advances in generating high-fidelity images.
Thus, in this work we propose a convolutional decoder for point clouds.
As shown by Groueix et al.~\cite{atlas18}, it is difficult to achieve high-quality auto-encoding results by training a na\"{i}ve point cloud decoder (i.e. a simple multi-layer perceptron).
In order to tackle this problem we propose several measures that allow for a better conditioning of the optimization problem.

Our contributions can be summarized as follows.
\begin{enumerate}
    \item We propose a convolutional decoder for point clouds that is able to outperform current state of the art results on autoencoding tasks.
    \item Our autoencoder is able to handle a varying number of points both for its input and output. This property makes it straight-forward to apply our architecture to the task of point cloud upsampling.
    \item To the best of our knowledge we are the first to apply Adaptive Instance Normalization as used in current image synthesis research \cite{adain18} to the area of point cloud generation. We give an intuition on why this technique is beneficial to training.
    \item We propose several additional losses to the commonly used Chamfer distance that consider both voxel-based and point cloud differences.
\end{enumerate}
Code and our sampling of the ShapeNet Core dataset (v2) \cite{chang2015shapenet} can be found at the project page \footnote{\href{https://graphics.rwth-aachen.de/publication/03303}{graphics.rwth-aachen.de/publication/03303}}.
\section{Related Work}
Most work on content synthesis with neural networks has been done on images.
The natural extension to 3D data is that of a voxel grid.
This allows a straightforward transfer of many image based methods (e.g.~by replacing 2D with 3D convolutions).
Examples are methods that deal with tasks such as 
single image shape reconstruction \cite{choy20163d}, shape completion \cite{han2017high}, and shape generation \cite{wu2016learning,li2017grass}.
Another option is to represent geometry as planar patches inserted into an Octree \cite{Wang-2018-AOCNN}.
However, as we are interested in point cloud methods we will restrict our discussion of related work to this domain.

\paragraph*{Point-Based Encoders}
Voxel-based approaches have its drawbacks when it comes to memory consumption, as the required memory scales cubically with the resolution of the grid.
To deal with these problems several architectures have emerged, that give up the regular grid structure and instead work directly on unordered point clouds. PointNet \cite{qi2017pointnet} is one of the first among those approaches and does not take any structure or neighbourhood into account. The internal shape representation here is created by aggregating point descriptors. 
As the relation between nearby points is often important to characterize shape, this work has been extended in PointNet++ \cite{qi2017pointnet++} where points are hierarchically grouped based on their neighbourhood and PointNet is applied to those local point clouds.
On the other hand, dynamic graph CNNs \cite{dgcnn} encode the information of a local neighborhood via graph convolutions.
PCNNs \cite{atzmon2018pcnn} generalize convolutions 
over points via the extension of the convolution operation to continuous volumetric functions.
In this manner they benefit from translational invariance and parameter sharing of convolutions, without the drawback of the memory size of high resolution voxel grids. 
Rethage et al.~\cite{fullycnn18} propose to combine the advantages of point clouds and grid structures by extracting features from points in the local neighbourhood of each grid cell using a network similar to PointNet.
On the resulting representation, 3D convolutions can be applied.
As a single grid cell encodes details of the point cloud and not just a binary occupancy value, a low resolution grid is sufficient.
This approach is most similar to the encoder used in our framework.

\paragraph*{Point Set Generation}
Most current approaches \cite{Fan2017APS,nash2017shape,Achlioptas2018LearningRA,mrt18,sonet18} for the generation of point clouds employ fully connected layers, sometimes in combination with upsampling and convolution layers, to generate a fixed number of points.
\cite{sonet18} employ both a convolution branch to recover the coarse shape and a fully connected branch for the details of the object. Unlike our approach they propose 2D convolutions that result in images with 3 channels, which are interpreted as point coordinates.
A different approach is taken by \cite{sinha2017surfnet} where instead of learning to output points directly, Sinha et al.~propose to learn a mapping from 2D to 3D.
By sampling the 2D domain one can obtain a point cloud.
This allows the number of generated points to be flexible.
Groueix et al.~\cite{atlas18} propose a method that builds on this approach.
However, instead of a single mapping a whole atlas of those is learned by training several networks in the style of \cite{sinha2017surfnet} that do not share parameters.
The loss then ensures that each network learns a different mapping and is responsible for a different part of the shape.
We employ a similar point generation technique in the sense that we also learn a mapping from 2D to 3D, instead of using fully connected layers to directly generate a fixed number of points.
However, we arrive at these maps in a different manner by generating them per grid cell with our proposed convolutional decoder. A different class of networks recently emerged to represent 3D shapes as an implicit function \cite{park2019deepsdf, chen2019implicit_decoder, mescheder2019occ}. This function can then be sampled to reconstruct explicit geometry.
\section{Convolutional Auto-Encoder for Point Clouds}
\begin{figure*}[tb]
    \centering
    \begin{overpic}[width=\textwidth]{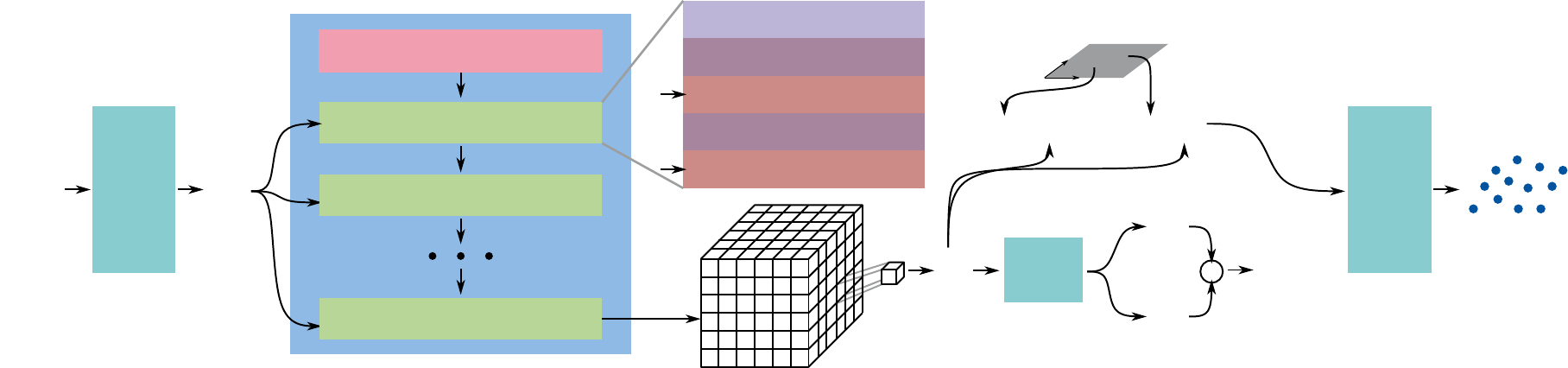}
    \put(2.5,11){$z$}
    \put(14,11){$w$}
    \put(6.75,11){MLP}
    \put(28.75,19.75){$P$}
    \put(28.75,15){$B$}
    \put(28.75,10.5){$B$}
    \put(28.75,2.5){$B$}
    \put(27,-1){decoder}
    \put(60,6){$f_c$}
    \put(56.5,4){$c$}
    \put(64.75,5.5){MLP}
    \put(86.75,10.75){MLP}
    \put(95,8){$\mathcal{Y}_c$}
    \put(73.5,8.5){$p_c$}
    \put(73.5,2.75){$\delta_c$}
    \put(81,6){$m$}
    \put(63,15){$\left[\epsilon_m,f_c\right]$}
    \put(72,15){$\left[\epsilon_1,f_c\right]$}
    \put(69,15){$\cdots$}
    \put(48,21.75){upsample}
    \put(47,19.25){convolution}
    \put(49,16.75){AdaIN}
    \put(47,14.5){convolution}
    \put(49,12){AdaIN}
    \put(40.25,18){$s_i, t_i$}
    \put(40.25,13.6){$s_j, t_j$}
    \end{overpic}
    \caption{Overview over our convolutional decoder:
    Given is some latent vector $z$ produced by an encoder.
    Passing it through a multi-layer perceptron (MLP) produces $w$, which consists of a series of scaling and translation parameters $\left[\left(s_1,t_1\right),\dots,\left(s_l,t_l\right)\right]$.
    $P$ is a learned constant parameter block (in our case it has dimension $512 \times 2 \times 2 \times 2$) used to kickstart the convolutional decoding process.
    The $B$ blocks each contain an upsampling layer (trilinear by a factor of $2$), followed by two convolution and AdaIN layers.
    The scaling and translation parameters from $w$ are used for each of the $l$ AdaIN layers in the convolutional decoder.
    The result is a voxel grid where each cell $c$ has a feature vector $f_c$.
    Using $f_c$ as input to a MLP we compute the probabilty $p_c$ that $c$ contains any point and the estimated local point cloud density $\delta_c$, which are then used together with the required output size $n$ to determine the number of points $m$ that should be generated for $c$.
    We then sample a uniform $2$-dimensional distribution (grey plane) $m$ times to produce $\epsilon_1,\dots,\epsilon_m$.
    Each $\epsilon_i$ is concatenated with $f_c$ as input to a MLP which produces a $3$-dimensional point. 
    Evaluating the MLP $m$ times produces a point cloud $\mathcal{Y}_c$ for grid cell $c$.
    }
    \label{fig:adain}
\end{figure*}
We want to represent our geometry as point clouds since they can approximate 3D shapes at a higher resolution without incurring the memory costs that voxel grids entail.
However, we also want to benefit from the advantages of grid structures, enabling the use of convolutional layers and Adaptive Instance Normalization (AdaIN).
To this purpose we propose our convolutional decoder (Section \ref{sec:decoder}), which starts out with a low resolution grid and successively increases the resolution up to the final desired grid size.
We then generate points for each grid cell.
Conversely, for our encoder (Section \ref{sec:encoder}) we embed the input point cloud into a voxel grid.
A network then encodes and stores local parts of the point cloud for each corresponding (closest) grid cell.
This voxel grid can then be encoded with a 3D convolutional network.

In traditional convolutional autoencoders the output of the encoder is passed to the decoder, 
who repeatedly upsamples it in order to produce the reconstruction of the input.
This means that even the encoding of fine details of the shape has to pass through the entire decoder, since high- and low-level features are not distinguished.
In contrast, our proposed decoder inserts the encoded shape information at various stages of the upsampling process.
We will explain our decoder in detail first, followed by the encoder.
In order to achieve high-quality results we introduce several additional losses.
\subsection{Decoder}
\label{sec:decoder}
Inspired by Karras et al.~\cite{adain18} we propose a convolutional decoder for point clouds based on Adaptive Instance Normalization (AdaIN) as used in a number of style-transfer methods \cite{dumoulin2017learned, ghiasi2017exploring, huang2017arbitrary, dumoulin2018feature}.
Given is an encoder that maps an input point cloud $\mathcal{X} \in \mathbb{R}^{n \times 3}$ to a latent vector $z \in \mathbb{R}^{1024}$.
A na\"{i}ve decoder would map $z$ to $\mathcal{Y} \in \mathbb{R}^{m \times 3}$ via a multi-layer perceptron (MLP). One problem with this approach is that a series of fully connected layers means adding a large number of parameters to the network.

Another problem is that in order to reconstruct fine detail of $\mathcal{X}$ in $\mathcal{Y}$ every layer of the network is required to preserve the entire shape information.
A small change in one of the parameters during back-propagation can have wide-reaching global effects on $\mathcal{Y}$.
While, one can reduce the number of parameters used by introducing a convolutional decoder, the problem of the interplay of different parameters during back-propagation remains.
Karras et al.~\cite{adain18} show that using AdaIN with a convolutional decoder/generator can produce impressive results for images.
An AdaIN layer works by first normalizing its input features and then applying an affine transformation per instance. The transformation parameters are an additional input (e.g. computed from $z$).
In practise, this means that our decoder is constructed via a series of upsampling, convolution, instance normalization \cite{instancenorm} and affine feature transformation layers followed by a nonlinearity (see Figure \ref{fig:adain}).
In contrast to traditional convolutional networks, the entire shape specific information is introduced through the affine transformations and is not passed through all layers of the decoder. Instead the upsampling process is applied to a learnable parameter block $P$.
For more details on the architecture see Appendix \ref{sec:arch}.

Thus a given $z$ (by some encoder) is mapped to a vector $w$ that contains the scaling and translation coefficients for each affine feature transformation layer.
For every layer $i$ with feature dimension $d$ where AdaIN is applied we select a slice $w_i \in \mathbb{R}^{2d}$.
We interpret $w_i = [s_i; t_i]$ such that $s_i, t_i \in \mathbb{R}^{d}$. As we regard only a single layer, we omit $i$ in the following.
The intermediate features $x = x^{(1)} \ldots\, x^{(d)}$ are first normalized and then scaled and translated:
\begin{equation}
\hat{x}^{(k)} = \frac{x^{(k)} - \mu(x^{(k)})}{\sqrt{\sigma^2(x^{(k)})+\epsilon}} \cdot s^{(k)} + t^{(k)},
\end{equation}
where $\mu(x^{(k)})$ and $\sigma^2(x^{(k)})$ are the mean and variance of $x^{(k)}$ over one instance.
Since all operations are done for each channel separately, in the following we will omit $k$ for readability.

As a result of this localized interaction the optimization problem becomes more well behaved.
Let $\nabla_{\hat{x}}\mathcal{L}$ be the gradient of a loss function (see Section \ref{subsec:loss}) with respect to the output of an intermediate normalization layer.
The gradient w.r.t.~a single cell $i$ of its input $x$ is given as
\begin{align}
    \nabla_{x_i}\mathcal{L} &= \frac{s}{\sqrt{\sigma^2(x)+\epsilon}}\left (\nabla_{\hat{x}_i}\mathcal{L} - \frac{\mathbb{1}^\intercal\left(\nabla_{\hat{x}}\mathcal{L}\right)}{m}- \frac{\hat{x}_i\left(\nabla_{\hat{x}}\mathcal{L}\right)^\intercal \hat{x}}{m} \right),
\end{align}
where $\hat{x} \in \mathbb{R}^m$ and $m$ is the number of cells.
For a scaling $a \in \mathbb{R}$ and a constant translation $b\cdot\mathbb{1} \in \mathbb{R}^m$, consider the case where $\nabla_{\hat{x}}\mathcal{L} = a \cdot \hat{x} + b\cdot\mathbb{1}$.
Then because $\hat{x}$ has zero mean and unit variance
\begin{align}
    \nabla_{x_i}\mathcal{L} &= \frac{s}{\sqrt{\sigma^2(x) + \epsilon}}\left (a \cdot \hat{x_i} + b - b - a \cdot \hat{x}_i \right) \\
    &= 0. \nonumber
\end{align}
Thus, there is no gradient w.r.t.~a scaling and translation of $x$ running through the normalization layer, which is only natural as such a transformation would be cancelled out by the normalization anyways.
AdaIN allows us to set this affine transformation individually for each object.
Therefore, the gradient w.r.t.~its parameters does not have to pass through the entire decoder. 
Consequently, the convolutional layers only have to learn non-affine interactions.

\subsubsection{Point Cloud Generation}
\label{sec:generation}
Our proposed convolutional decoder so far only generates volumetric grids. We are however interested in generating point clouds.
Therefore, as shown in Figure \ref{fig:adain}, for each cell $c$ we feed its encoded information $f_c$ into a simple MLP.
This MLP predicts two values $p_c$ and $\delta_c$. $p_c$ is a binary variable predicting whether a cell is filled or empty. 
$\delta_c$ is a probability density function (i.e. the likelihood, that a sample should be generated for a particular cell).
Distributing the estimation of this information over two variables helps us in dealing with empty cells, as the density prediction seldom actually reaches zero and we thus would introduce points at unwanted locations.
For all cells that are classified as filled we then distribute the total number of output samples proportionally to the density estimates of the cells. Thus our network is independent of the number of points we want to generate. This number can even be changed between training and inference (see Figure \ref{fig:teaser}).

The actual generation of points is done in a similar manner to Groueix et al.~\cite{atlas18} and Yin et al.~\cite{yin2018p2pnet}.
The idea is to learn a parameterization from a $k$-dimensional domain to $\mathbb{R}^3$. Then by randomly sampling this domain from a uniform distribution and applying the map, we get our $3$-dimensional points. 
In all our experiments we set $k=2$, since we assume that locally the shape can be approximated with a surface patch.
In practice we apply this map by concatenating the $k$-dimensional sample to the encoded cell information $f_c$ and feeding
the resulting vector into a MLP, which outputs a $3$-dimensional point.
Thus the MLP represents a map $m_{f_c}: \mathbb{R}^{k} \to \mathbb{R}^3$ conditioned on $f_c$.
During inference we sample the $k$-dimensional domain uniformly and then apply a number of steps of Lloyd's algorithm \cite{lloyd1982least} to ensure an even coverage of the space.
This further improves our results as shown in Section \ref{sec:ablation}. The predicted samples of each cell are offset by the corresponding cell centers.

\subsection{Encoder}
\label{sec:encoder}
\begin{figure}
    \centering
    \begin{overpic}[width=\columnwidth]{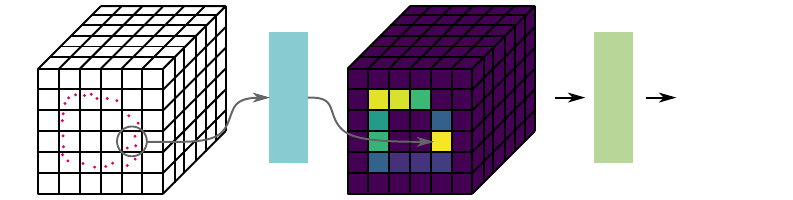}
    \put(86,12){$z \in \mathbb{R}^{1024}$}
    \put(35,6){\rotatebox{90}{PointNet}}
    \put(76,6){\rotatebox{90}{3D CNN}}
    \end{overpic}
    \caption{Our convolutional encoder follows a similar method to Rethage et al.~\cite{fullycnn18}. We embed the input point cloud (red) into a volumetric grid. For each cell we pass all points within a certain radius from the cell center into a small PointNet. This results in a grid where each cell encodes local point cloud information via a $32$-dimensional feature vector. This is visualized as a multi-colored grid. The grid is then passed through a 3D CNN. Through a series of convolution and max-pooling layers we compute an encoding $z \in \mathbb{R}^{1024}$ of the input point cloud.}
    \label{fig:encoder}
\end{figure}
For our encoder (see Figure \ref{fig:encoder}) we follow a similar approach to Rethage et al.~\cite{fullycnn18}.
We isotropically normalize the input point cloud such that the longest edge of its axis-aligned bounding box is scaled to the range $\left[-0.5, 0.5\right]$.
This point cloud $\mathcal{X}$ is then embedded into a volumetric grid consisting of $32^3$ cells.
For each grid cell we encode the local neighborhood of $\mathcal{X}$ (all points within a radius $r = \frac{\sqrt{3}}{2}$ to the cell center) via a small PointNet (proposed by Qi et al.~\cite{qi2017pointnet}).
Apart from using fewer number of parameters we also aggregate the final encoding of point clouds by computing the mean of the point features instead of the maximum as proposed in the original paper.
Since we make use of a PointNet we are able to handle input point clouds with varying number of points.

This results in a grid where each cell has an $\eta$-dimensional feature vector ($\eta=32$ in all our experiments).
This grid can then be passed through a 3D CNN, which consists of a series of convolution, batchnorm and max-pooling layers.
The output is an encoding $z \in \mathbb{R}^{1024}$ of $\mathcal{X}$.
For more details on the architecture see Appendix \ref{sec:arch}.
\subsection{Loss Functions}
\label{subsec:loss}
We define the distance of a point $s_{\mathcal{X}}$ to a point cloud $\mathcal{Y}$ as
\begin{align}
d(s_{\mathcal{X}}, \mathcal{Y}) = \min_{s_{\mathcal{Y}} \in \mathcal{Y}} \norm{s_{\mathcal{X}} - s_{\mathcal{Y}}}_2.
\end{align}
In order to compare the input point cloud $\mathcal{X} \in \mathbb{R}^{n \times 3}$ to the reconstructed point cloud $\mathcal{Y} \in \mathbb{R}^{m \times 3}$ we measure the difference with the commonly used Chamfer distance as proposed for point clouds in \cite{Fan2017APS},
\begin{align}
    L_{c}(\mathcal{X},\mathcal{Y}) 
    = \frac{1}{n} \sum_{s_{\mathcal{X}} \in \mathcal{X}} d(s_{\mathcal{X}}, \mathcal{Y})^2 
    + \frac{1}{m} \sum_{s_{\mathcal{Y}} \in \mathcal{Y}} d(s_{\mathcal{Y}}, \mathcal{X})^2.
\end{align}
This gives us a gradient for every point in $\mathcal{Y}$.
However, we found that additionally formulating a sharper version of the Chamfer distance benefits training (see Section \ref{sec:res}).
With the formulation
\begin{align}
    L_{p}(\mathcal{X},\mathcal{Y}) 
    = \frac{1}{n} \sqrt[\leftroot{0}\uproot{8}p]{\sum_{s_{\mathcal{X}} \in \mathcal{X}} d(s_{\mathcal{X}}, \mathcal{Y})^p}
    + \frac{1}{m} \sqrt[\leftroot{0}\uproot{8}p]{\sum_{s_{\mathcal{Y}} \in \mathcal{Y}} d(s_{\mathcal{Y}}, \mathcal{X})^p}.
\end{align}
the gradients of points that incur a larger error are weighted more heavily with $p > 2$. For high $p$ this measure can be seen as similar to the Hausdorff distance. In our experiments we used $p = 5$.

Since $\mathcal{Y}$ is generated by offsetting generated per-cell point clouds $\mathcal{Y}_c$ by the corresponding cell centers $c_o$, we want to enforce a notion of locality (i.e. each cell only contributes to the part of $\mathcal{Y}$ in its vicinity).
Thus we add a loss
\begin{align}
L_{o}(\mathcal{Y}) = \sum_c \sum_{s_c \in \mathcal{Y}_c} \max \left(\,\text{dist}(s_c,c_o) - m,\, 0\,\right),
\end{align}
This penalizes any generated points that are too far away from their cell centers. We choose $m = \sqrt{3}$ to allow points to be distributed within their generating cell and its direct neighbours.

We cannot directly train the density estimates and filled cell predictions using only the point-wise differences shown above.
This is because the differences do not give a gradient w.r.t. the number of points per cell.
For this reason we generate ground truth densities and label the filled cells based on the input.
Training the MLP that predicts the density $\delta$ and probability that a cell $c$ is filled $p$ is done by using the mean squared error
\begin{align}
    L_{d}(\delta, \hat{\delta}) = \frac{1}{32^3} \sum_c  \left (\delta_c - \hat{\delta}_c \right)^2,
\end{align}
and the binary cross entropy loss
\begin{align}
    L_{f}(p, \hat{p}) = -\frac{1}{32^3} \sum_c \hat{p}_c\cdot\log(p_c) + (1-\hat{p}_c) \cdot \log(1 - p_c)
\end{align}
respectively.
Here $\hat{\delta}$ and $\hat{p}$ denote the ground truth.
Thus our loss during training is
\begin{align}
\lambda_1 L_c(\mathcal{X},\mathcal{Y}) &+ \lambda_2 L_p(\mathcal{X},\mathcal{Y}) + \lambda_3 L_d(\delta,\hat{\delta})\\
&+ \lambda_4 L_f(p, \hat{p}) + \lambda_5 L_o(\mathcal{Y}) \nonumber
\end{align}
In all our experiments we chose ${\lambda_1 = \num{1e3}}$, ${\lambda_2 = \num{1e1}}$, ${\lambda_3 = \num{1e10}}$, ${\lambda_4 = \num{1e2}}$, and ${\lambda_5 = 1}$.
\section{Experiments}
\label{sec:res}
We evaluate our decoder network both by showing the effectiveness of several design choices and by comparing our results with the current state of the art on the task of autoencoding 3D point clouds. 
All our experiments with our proposed method were done on the ShapeNet dataset \cite{chang2015shapenet}, where we evaluated both our method and the methods proposed in \cite{atlas18, sonet18}.
Additionally, we performed experiments using their respective settings and datasets.
This is necessary for a thorough comparison, since prior work employs different datasets, data normalization techniques and evaluation criteria.
Furthermore, we can assume that their proposed network architectures were tuned according to the respective datasets.
Our networks were trained using AMSGrad~\cite{j.2018on} ($\beta_1 = 0.9$, $\beta_2 = 0.999$, learning rate $= 0.0046$).
For evaluation on the testing set we used the network weights that performed best on the validation set.
All other networks were trained using the hyper-parameters settings suggested in the respective works.

\paragraph*{Dataset}
For our experiments we made use of the official training, validation, and testing split of the ShapeNet Core dataset (v2), which consists of ca.~50k models in 55 different categories.
We found that a high quality sampling is important to achieve good results (see Table \ref{tab:ablation}), as the loss is strongly affected by it.
Minimizing the Chamfer distance on a non-even sparse sampling does not necessarily mean that we are able to achieve a good approximation of the underlying surface.
A large distance from a reconstructed point to the closest target point can be either caused by a great distance to the underlying surface (which we want to penalize) or by the lack of samples in this particular part of the surface (which we do not want to penalize).
Therefore, it is desirable that the sampling is as even as possible over the entire shape.
To achieve such a sampling, we strongly oversampled the objects uniformly (with roughly 80k points) and then chose a subset (16k points) of those with farthest point sampling.

As our encoder sorts all points into a grid, we normalize the point clouds to the size of the unit cube centered in the origin. No further data augmentation is applied. All metrics are however computed on unnormalized shapes to simplify future comparisons. When not mentioned otherwise, all distances are reported between point clouds with 2500 points.

\subsection{Ablation Study}

\begin{figure*}
    \centering
    \begin{overpic}[height=\textheight - 36pt]{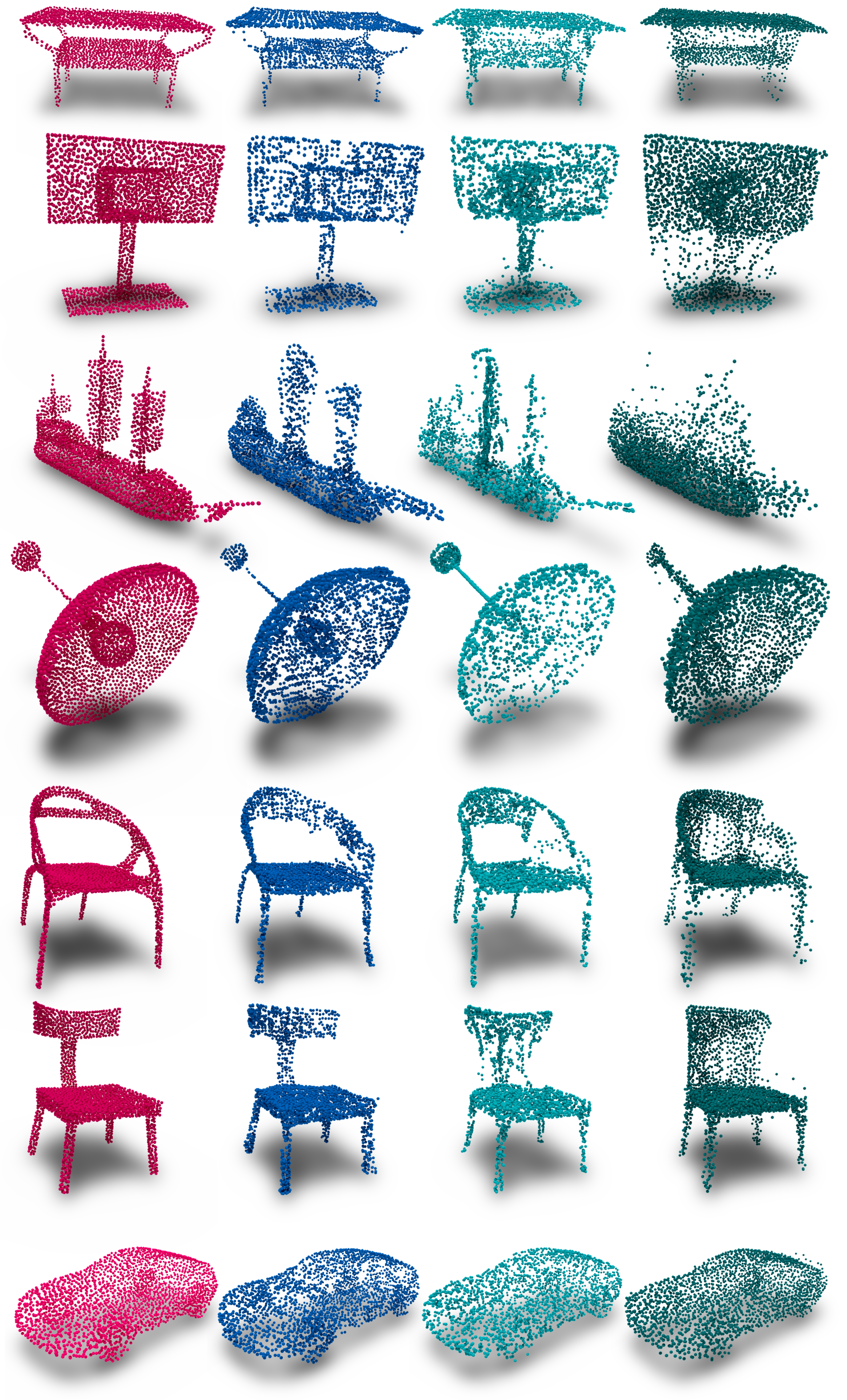}
    \put(7,100){Input}
    \put(20,100){our method}
    \put(31,100){AtlasNet (125 patches)}
    \put(50,100){SO-Net}
    \end{overpic}
    \caption{Qualitive results for different autoencoder models. From left to right: ground truth, our results, \cite{atlas18}, \cite{sonet18}. Note that our method produces less spurious points and reproduces sharper surface details.}
    \label{fig:examples}
\end{figure*}

\label{sec:ablation}
\begin{table}[tb]
    \centering
    \begin{tabular}{||l|c||}
        \hline
         method & Chamfer dist. \\
         \hline \hline
         (1) with randomly sampled point clouds & $0.387$\\
         (2) without AdaIN & $0.385$ \\
         (3) without regularization loss & $0.401$\\
         (4) without p-norm & $0.384$\\
         (5) all of the above & $0.440$ \\
         (6) with randomly sampled map & $0.390$\\
         (7) our method (9 transformations) & $0.401$\\
         (8) our method (3 transformations) & $\mathbf{0.376}$ \\
         \hline
         (9) random sampling & $0.227$ \\
         \hline
    \end{tabular}
    \caption{Evaluation of the different design choices for our network. As can be seen, each additional loss, sampling and architecture choice improves the final result (bold). The reported metric is the Chamfer distance as introduced in section \ref{subsec:loss} multiplied with 1000. To put the numbers into context we compare a random sampling of the same shape with the target.}
    \label{tab:ablation}
\end{table}

To motivate our design choices we performed an extensive ablation study, reporting the Chamfer distance obtained on the testing set for different changes in our input, architecture or loss function (Table \ref{tab:ablation}). 
To show the effect of an evenly distributed point cloud, we trained the network on a uniform random sampling (1) as used in \cite{sonet18}. 
We evaluated on our high quality point clouds.
To motivate the use of AdaIN, we implemented a strong baseline in the form of a convolutional autoencoder.
We used the same encoder as in our proposed network.
However, for the decoder we used a convolutional decoder without AdaIN (2) (i.e. $z$ is passed directly into the decoder and $P$ is no longer necessary).
To ensure a fair comparison we used a similar number of parameters.

While our proposed architecture enables the possible application of nine layers of AdaIN (7), we found that this lead to some overfitting on the training data.
Therefore, we limit the number of affine feature transformations to the first three layers (8).
All subsequent outputs of instance normalization layers are not scaled and translated.
This architecture achieved the best result (marked in bold in Table \ref{tab:ablation}).

To show the effectiveness of the additionally introduced losses, we trained networks without them and show the difference in the resulting Chamfer distance (3,4).
For further comparison, we trained a network in a fairly simple manner by only using the chamfer distance as a loss and no AdaIN on randomly sampled point clouds (5).
Finally, we show that sampling the learned map from 2D to 3D at fixed, well distributed positions (as done in \cite{atlas18}) instead of randomly during inference further improves the results (6).
Not using the cell classification loss has a minor negative impact on the results in the order of the fourth decimal. 
To put these numbers into context, we compare a random sampling of the shape with the ground truth (9).

\subsection{Comparison}
We compare against AtlasNet \cite{atlas18} and SO-Net \cite{sonet18} both on our own dataset (Table \ref{tab:our_data}) as well as on their respective datasets (Table \ref{tab:their_data}).
For AtlasNet we trained their best performing network (125 Patches) on our dataset.
SO-Net does not allow to output point clouds with 2500 points without changing the suggested architecture.
Instead, we compare against the two presented versions of the network.
One generates 1280 points (Table \ref{tab:their_data}) and one has an output size of 4608 points (Table \ref{tab:our_data}).
The numbers reported in their paper are from a network outputting 1280 points, consequently we trained ours similarly (i.e. 1024 input points and 1280 output points).
Furthermore, they use a slightly different definition of the Chamfer distance.
They compute the Euclidean distance between closest points instead of its squared version. 
For a fair comparison on our dataset we report the Chamfer distance between a target of 2500 points and the entire point cloud (4608 points) as well as subsamplings (2500 points) of it. 
\begin{table}[tb]
    \centering
    \resizebox{\columnwidth}{!}{%
    \begin{tabular}{||l|c||}
        \hline
         method & Chamfer dist. \\
         \hline \hline
         SO-Net ($4608$ points) & $0.603$ \\
         SO-Net ($2500$ points via random subsampling) & $0.708$ \\
         SO-Net ($2500$ points via farthest point sampling) & $0.691$ \\
         AtlasNet ($125$ patches) & $0.408$ \\
         our method & $\mathbf{0.376}$\\
         \hline
    \end{tabular}
    }
    \caption{Comparison of our method against SO-Net \cite{sonet18} and AtlasNet \cite{atlas18} on our dataset. The reported number is the Chamfer distance multiplied by 1000.}
    \label{tab:our_data}
\end{table}

Note that the computed distances are not comparable across datasets due to differences in normalization and evaluation methods.
As can be seen in Tables \ref{tab:our_data} and \ref{tab:their_data} our method outperforms AtlasNet and SO-Net on our dataset as well as on the ones used by the respective authors. Qualitative results are shown in Figure~\ref{fig:examples}.
For these examples, our method is less prone to produce outliers and reconstructs the shape contours more faithfully.
\begin{table}[tb]
    \centering
    \begin{tabular}{||l|c||}
        \hline
         method & Chamfer dist. \\
         \hline \hline
         AtlasNet ($25$ Patches) & $1.56$\\
         AtlasNet ($125$ Patches) & $1.51$\\
         our method &  $\mathbf{1.42}$\\
         \hline \hline
         SO-Net ($1280$ points) & $0.033$ \\
         our method ($1280$ points) & $\mathbf{0.030}$ \\
         \hline
    \end{tabular}
    \caption{Comparison against AtlasNet and SO-Net on their respective datasets. Our models were trained without any additional hyper-parameter tuning. The reported number for the comparison against AtlasNet is the Chamfer distance multiplied by 1000. The comparison against SO-Net is based on the Chamfer distance as reported in their paper \cite{sonet18}.}
    \label{tab:their_data}
\end{table}
\section{Applications}
\label{sec:app}
To demonstrate the usefulness of our convolutional decoder we show results in three applications.
Our hyper-parameters and architecture were not tuned particularly for these demonstrations.
We expect that with more carefully chosen settings, better results could be achieved.
\paragraph*{Single View Reconstruction}
\begin{figure}[tb]
    \centering
    \includegraphics[width=\columnwidth]{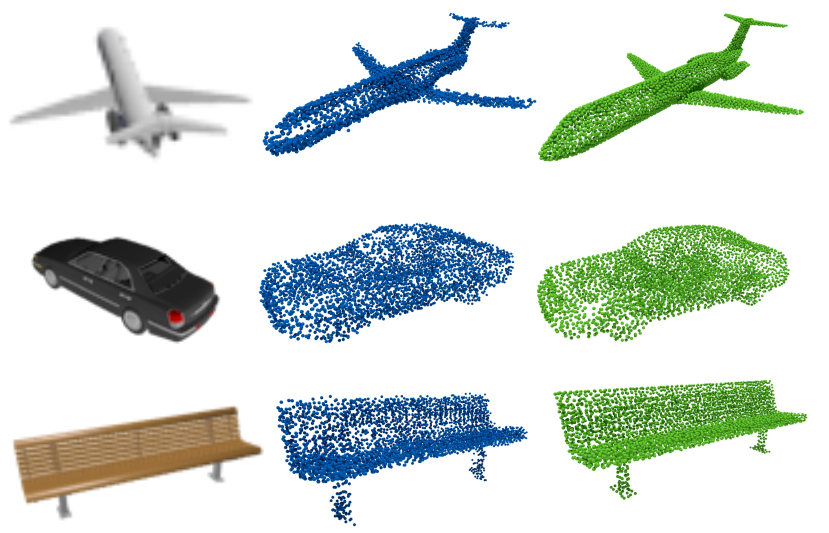}
    \caption{We show some qualitative results for single view reconstruction. The input images are shown on the left. Reconstruction results are visualized in blue. The ground truth is rendered in green.}
    \label{fig:svr}
\end{figure}
For single view reconstruction (see Figure \ref{fig:svr}) we follow \cite{choy20163d} and use a subset of ShapeNet consisting of 13 classes.
To be comparable we use their rendered views, as well as their sampling.
Similar to \cite{mrt18} we used a pretrained VGG-11 \cite{simonyan2014very} as an encoder. The rest of our network is unchanged to the autoencoder setting.
We manage to achieve competitive quantitative results as shown in Table \ref{tab:svr}.
\begin{table}[ht]
    \centering
    \begin{tabular}{||l|c||}
        \hline
         method & Chamfer dist. \\
         \hline \hline
         Fan et al. \cite{Fan2017APS}  & $4.128$ \\
         Lin et al. \cite{lin2018learning} & $3.547$ \\
         MRTNet \cite{mrt18} & $\mathbf{3.088}$ \\
         our method & $3.398$ \\
         \hline
    \end{tabular}
    \caption{Quantitative results for Single View Reconstruction. The reported numbers are Chamfer distance (as defined in \cite{sonet18}), scaled by 100, computed on point clouds of size 4096}
    \label{tab:svr}
\end{table}

\paragraph*{Point Cloud Upsampling}
\begin{figure}[tb]
    \centering
    \includegraphics[width=\columnwidth]{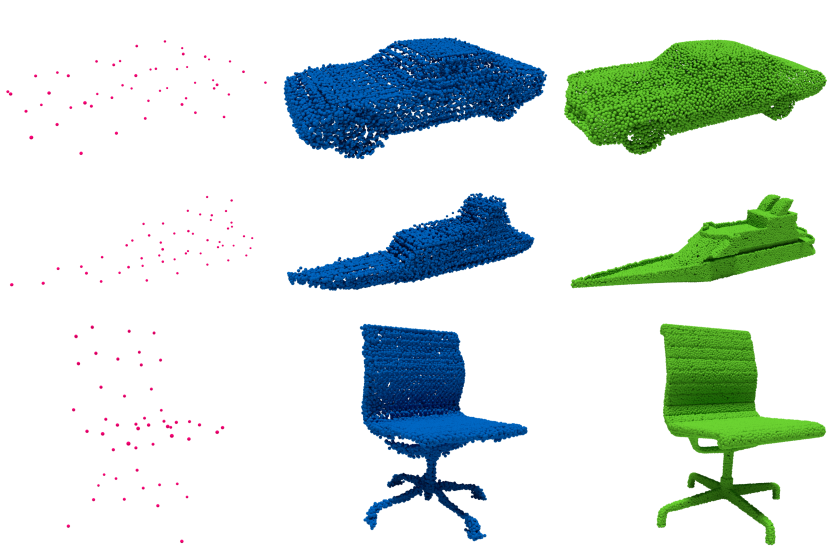}
    \caption{Qualitative results for our point cloud upsampling. Severely under-sampled input point clouds (50 points) are shown in red. The network predictions and ground truth point clouds are shown in blue and green respectively (16000 points).}
    \label{fig:upsampling}
\end{figure}
As our network architecture is indifferent to the number of input or output points, it is straightforward to use our model for the task of point cloud upsampling.
We train our network on our training set to take between 50 and 500 input points, but output 5000.
Although there are several methods that use neural networks for point cloud upsampling \cite{yifan2018patch,yu2018pu}, their setting is different as they regard local patches of the geometry and compute a denser sampling there.
In contrast, we regard the shape as a whole.
As a result these methods require the input to be sampled densely enough that local patches convey geometric meaning.
For our method it is sufficient that the general shape is conveyed in order to get results of a good quality. We demonstrate this on severely under-sampled point clouds of the test set with only 50 points as input (Figure \ref{fig:upsampling}).
Note that our method is able to robustly output point clouds of size 16000 even though the network was trained to output 5000 points.

\begin{figure}[tb]
    \centering
    \includegraphics[width=\columnwidth]{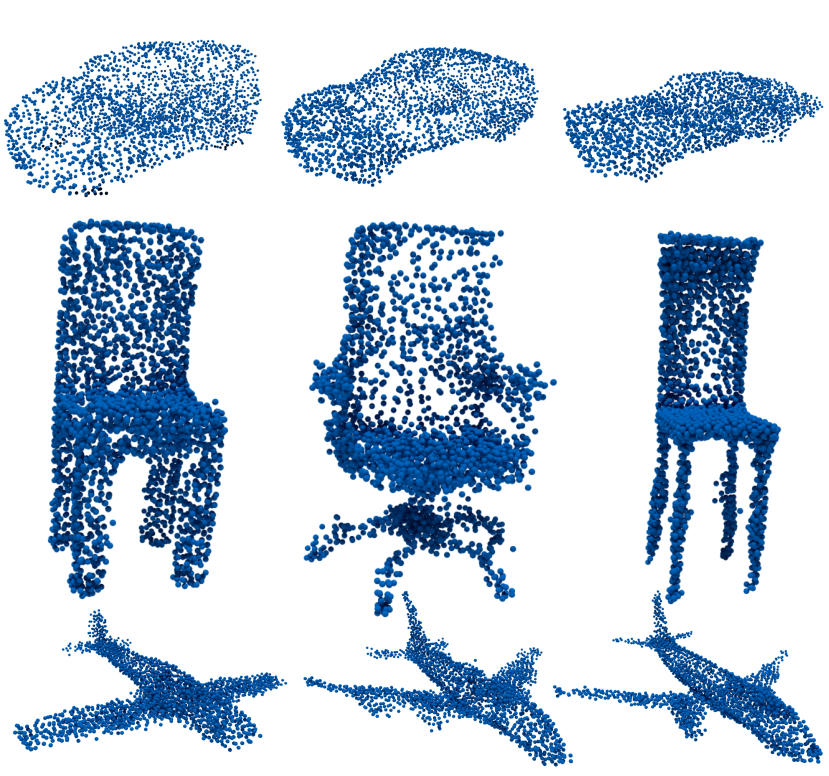}
    \caption{Here we show some samples generated with our conditional VAE for the categories "car", "chair", and "airplane".}
    \label{fig:vae}
\end{figure}

\paragraph*{Point Cloud Synthesis}
Our decoder can not only be used to reconstruct point clouds for a given input but is also able to generate new shapes as well. 
A commonly used generative model is the variational autoencoder (VAE) as proposed by Kingma et al.~\cite{kingma2013auto}.
We implemented a conditional VAE version of our network, with only minor changes to the original autoencoder.
Conditioning on different classes is done by passing the category as a one-hot encoding vector into a MLP, which generates $P$ (see Figure \ref{fig:adain}).
The latent vector $z$ is sampled from a multivariate Gaussian, whose parameters are predicted by the encoder.
This allows us to sample the latent space in order to generate shapes for a specified category as shown in Figure \ref{fig:vae}.
\section{Conclusion}
In this work we have introduced a convolutional decoder that can generate high quality point clouds of arbitrary size.
Our method is able to achieve state of the art results for auto-encoding tasks by making use of the benefits offered by AdaIN, careful consideration of even sampling, as well as several additions to the Chamfer distance as losses.
We outline several possible applications for our method in the fields of single view reconstruction, point cloud upsampling and synthesis.

Our architecture inherits some of the common limitations that come with voxel-based representations. That is, our method is not invariant to rotations of the input and could incur a larger memory cost at higher grid resolutions. However, we show that with a relatively low resolution ($32^3$) we are able to generate results of a high quality.
Furthermore, we approximate the geometry in each filled grid cell as a surface patch.
For locally more complex geometries this might be a limitation.

Nevertheless, we are convinced that our method is useful in future research on 3D shape synthesis.
One direction is the use of a generator similar to our decoder in the setting of generative adversial networks (GANs) as originally proposed by Goodfellow et al.~\cite{goodfellow2014generative}.
Another interesting research direction are more detailed shape modifications enabled by affine feature transformations at varying levels of detail.
\paragraph*{Acknowledgements}The research leading to these results has received funding from the European Research Council under the European Union's Seventh Framework Programme
(FP7/2007-2013)/ERC grant agreement n$^\circ$ [340884], as well as the Deutsche Forschungsgemeinschaft DFG -- 392037563.
\appendix
\section{Network Architecture}
\label{sec:arch}
Our encoder consists of a small PointNet and an 3D CNN.
The PointNet is constructed as FC8-FC16-FC32-FC32.
FC$x$ is a fully connected layer (in this case without bias) with output dimensionality $x$.
After every fully connected layer we apply batchnorm as proposed by Ioffe and Szegedy \cite{ioffe2015batch}.
We also apply the exponential linear unit (ELU) as an activation function as proposed by Clevert et al.~\cite{clevert2015elu} after every batchnorm layer except for the last one.
In order to construct the final $32$-dimensional feature vector for each cell we compute the mean feature instead of taking the maximum.

The 3D CNN is constructed as C64-C64-C64-MP-C128-C128-MP-C256-C256-MP-C512-C512-MP-C512-C1024.
C$x$ is a 3D convolution layer with kernel size $3\times3\times3$, zero-padding of 1, stride of 1, and output feature dimensionality $x$.
For C1024 we use no padding and a kernel size of $2\times2\times2$ in order to reduce the output to a 1024-dimensional vector.
We do not use bias for our convolution operations.
After every convolution layer we apply batchnorm and ELU.
MP refers to a max-pooling layer with kernel size $2\times2\times2$ and stride 1.

For our decoder we use a fully connected layer with bias to map $z$ to $w$.
The convolutional decoder is constructed as P-C512-U-C512-C256-U-C256-C128-U-C128-C64-U-C64-C62.
P refers to the learnable constant parameter block of size $512\times2\times2\times2$.
C$x$ refers to 3D convolution layers with output feature dimensionality $x$, kernel size $3\times3\times3$, stride of 1, and zero-padding of 1.
We do not use bias for our convolution operations.
After every convolution and P we apply dropout as proposed by Srivastava et al.~\cite{dropout} with a probabilty of 0.2.
AdaIN is applied after every dropout layer and after P with the scaling and translation parameters provided by $w$.
For every convolution layer we apply ELU after AdaIN.

Our point cloud generation MLP is structured as FC64-FC64-FC32-FC32-FC16-FC16-FC8-FC3.
We apply ELU after every FC layer except for the last one.

The MLP that estimates the density and classifies whether a grid cell contains points or not is constructed as FC16-FC8-FC4-FC2.
After every fully connected layer we apply batchnorm and ELU except for the last one.

\bibliographystyle{eg-alpha-doi}  
\bibliography{main}        


\end{document}